\documentclass[10pt,twocolumn,letterpaper]{article}

\usepackage{tikz}
\usetikzlibrary{arrows.meta, positioning, calc}
\usepackage{cvpr}              

\usepackage{adjustbox}
\usepackage{multirow}
\usepackage{tabularx}
\usepackage{makecell}
\usepackage[accsupp]{axessibility} 

\definecolor{cvprblue}{rgb}{0.21,0.49,0.74}
\usepackage[pagebackref,breaklinks,colorlinks,citecolor=cvprblue]{hyperref}

\def\paperID{*****} 
\def\confName{CVPR}
\def\confYear{2025}

\title{The First Challenge on Remote Sensing Infrared Image Super-Resolution at NTIRE 2026: Benchmark Results and Method Overview}
\author{
Kai Liu$^{\dagger}$ \and
Haoyang Yue$^{\dagger}$ \and
Zeli Lin$^{\dagger}$ \and
Zheng Chen$^{\dagger}$ \and
Jingkai Wang$^{\dagger}$ \and
Jue Gong$^{\dagger}$ \and
Jiatong Li$^{\dagger}$ \and
Xianglong Yan$^{\dagger}$ \and
Libo Zhu$^{\dagger}$ \and
Jianze Li$^{\dagger}$ \and
Ziqing Zhang$^{\dagger}$ \and
Zihan Zhou$^{\dagger}$ \and
Xiaoyang Liu$^{\dagger}$ \and
Radu Timofte$^{\dagger}$ \and
Yulun Zhang$^{\dagger*}$ \and
Junye Chen \and
Zhenming Yan \and
Yucong Hong \and
Ruize Han \and
Song Wang \and
Li Pang \and
Heng Zhao \and
Xinqiao Wu \and
Deyu Meng \and
Xiangyong Cao \and
Weijun Yuan \and
Zhan Li \and
Zhanglu Chen \and
Boyang Yao \and
Yihang Chen \and
Yifan Deng \and
Zengyuan Zuo \and
Junjun Jiang \and
Saiprasad Meesiyawar \and
Sulocha Yatageri \and
Nikhil Akalwadi \and
Ramesh Ashok Tabib \and
Uma Mudenagudi \and
Jiachen Tu \and
Yaokun Shi \and
Guoyi Xu \and
Yaoxin Jiang \and
Cici Liu \and
Tongyao Mu \and
Qiong Cao \and
Yifan Wang \and
Kosuke Shigematsu \and
Hiroto 	Shirono \and
Asuka Shin\and
Wei Zhou \and
Linfeng Li \and
Lingdong Kong \and
Ce Wang \and
Xingwei Zhong \and
Wanjie Sun \and
Dafeng Zhang \and
Hongxin Lan \and
Qisheng Xu \and
Mingyue He \and
Hui Geng \and
Tianjiao Wan \and
Kele Xu \and
Changjian Wang \and
Antoine Carreaud \and
Nicola Santacroce \and
Shanci Li \and
Jan Skaloud \and
Adrien Gressin
}

\begin{document}

\maketitle

\let\thefootnote\relax\footnotetext{$^{\dagger}$Kai Liu, Haoyang Yue, Zeli Lin, Zheng Chen, Jingkai Wang, Jue Gong, Jiatong Li, Yanxiang Long, Libo Zhu, Jianze Li, Ziqing Zhang, Zihan Zhou, Xiaoyang Liu, and Radu Timofte, Yulun Zhang are the challenge organizers, while the other authors participated in the challenge. $^{*}$Corresponding author: Yulun Zhang. Section B in the supplementary materials contains the authors' teams and affiliations. NTIRE 2026 webpage: \url{https://cvlai.net/ntire/2026}. Code: \url{https://github.com/Kai-Liu001/NTIRE2026_infraredSR}.}

\vspace{-10.mm}
\begin{abstract}
This paper presents the NTIRE 2026 Remote Sensing Infrared Image Super-Resolution (×4) Challenge, one of the associated challenges of NTIRE 2026. The challenge aims to recover high-resolution (HR) infrared images from low-resolution (LR) inputs generated through bicubic downsampling with a ×4 scaling factor. The objective is to develop effective models or solutions that achieve state-of-the-art performance for infrared image SR in remote sensing scenarios. To reflect the characteristics of infrared data and practical application needs, the challenge adopts a single-track setting. A total of 115 participants registered for the competition, with 13 teams submitting valid entries. This report summarizes the challenge design, dataset, evaluation protocol, main results, and the representative methods of each team. The challenge serves as a benchmark to advance research in infrared image super-resolution and promote the development of effective solutions for real-world remote sensing applications.
\end{abstract}

\setlength{\abovedisplayskip}{1pt}
\setlength{\belowdisplayskip}{1pt}

\vspace{-6mm}
\section{Introduction}
\vspace{-2mm}
Infrared image super-resolution aims to enhance the spatial resolution of low-resolution infrared remote sensing images while preserving critical scene details and thermal radiometric consistency. Due to limitations in sensor hardware and imaging conditions, infrared images often suffer from low spatial resolution, which obscures fine-grained structures such as small targets, texture details, and subtle thermal variations. This degradation directly affects the reliability of downstream remote sensing tasks, including target detection, environmental monitoring, and change analysis.

Accurate infrared image super-resolution requires reconstructing high-resolution images that are both structurally consistent and thermally faithful to the original scene. 
This task is inherently challenging due to the presence of noise in infrared data, the non-linear relationship between thermal radiation and pixel intensity, and the need to preserve meaningful thermal contrasts across different land-cover types. In addition, variations in atmospheric conditions and sensor characteristics further complicate the generalization of SR models in real-world scenarios.

With the rapid development of deep learning, a wide range of methods have been explored for image super-resolution.
Convolutional neural networks (CNNs) have demonstrated strong capability in learning hierarchical representations for detail reconstruction~\cite{dong2014learning,kim2016accurate,zhang2018image}. Building upon this foundation, subsequent works incorporate deeper architectures, residual connections, and attention mechanisms to improve reconstruction accuracy~\cite{zhang2018residual,dai2019second}. More recently, Transformer-based models enhance performance by modeling long-range dependencies via self-attention~\cite{vaswani2017attention,liu2021swin,chen2023dual}, while state-space models (e.g., Mamba) provide efficient alternatives for large-scale modeling~\cite{gu2023mamba,liu2024vmamba,guo2024mambair}. In addition, generative approaches, including GAN-based methods~\cite{goodfellow2014generative,ledig2017photo} and diffusion models~\cite{ho2020denoising,saharia2022image,xia2023diffir,wu2023seesr,li2024distillation}, have shown strong capability in synthesizing realistic textures and improving reconstruction quality.

To provide a standardized benchmark for infrared image super-resolution, we organize the NTIRE 2026 Remote Sensing Infrared Image Super-Resolution (×4) Challenge, associated with the NTIRE 2026 Workshop. The challenge focuses on reconstructing high-resolution infrared images from low-resolution inputs generated via bicubic downsampling with a ×4 scaling factor.

This challenge adopts a single-track setting, where all submissions are evaluated using a unified image quality assessment (IQA) metric defined as PSNR + 20×SSIM. The evaluation is conducted on single-channel infrared images, consistent with the characteristics of the data.

This challenge is one of the challenges associated with the NTIRE 2026 Workshop~\footnote{\url{https://www.cvlai.net/ntire/2026/}} on:
deepfake detection~\cite{ntire26deepfake}, 
high-resolution depth~\cite{ntire26hrdepth},
multi-exposure image fusion~\cite{ntire26raim_fusion}, 
AI flash portrait~\cite{ntire26raim_portrait}, 
professional image quality assessment~\cite{ntire26raim_piqa},
light field super-resolution~\cite{ntire26lightsr},
3D content super-resolution~\cite{ntire263dsr},
bitstream-corrupted video restoration~\cite{ntire26videores},
X-AIGC quality assessment~\cite{ntire26XAIGCqa},
shadow removal~\cite{ntire26shadow},
ambient lighting normalization~\cite{ntire26lightnorm},
controllable Bokeh rendering~\cite{ntire26bokeh},
rip current detection and segmentation~\cite{ntire26ripdetseg},
low light image enhancement~\cite{ntire26llie},
high FPS video frame interpolation~\cite{ntire26highfps},
Night-time dehazing~\cite{ntire26nthaze,ntire26nthaze_rep},
learned ISP with unpaired data~\cite{ntire26isp},
short-form UGC video restoration~\cite{ntire26ugcvideo},
raindrop removal for dual-focused images~\cite{ntire26dual_focus},
image super-resolution (x4)~\cite{ntire26srx4},
photography retouching transfer~\cite{ntire26retouching},
mobile real-word super-resolution~\cite{ntire26rwsr},
remote sensing infrared super-resolution~\cite{ntire26rsirsr},
AI-Generated image detection~\cite{ntire26aigendet},
cross-domain few-shot object detection~\cite{ntire26cdfsod},
financial receipt restoration and reasoning~\cite{ntire26finrec},
real-world face restoration~\cite{ntire26faceres},
reflection removal~\cite{ntire26reflection},
anomaly detection of face enhancement~\cite{ntire26anomalydet},
video saliency prediction~\cite{ntire26videosal},
efficient super-resolution~\cite{ntire26effsr},
3d restoration and reconstruction in adverse conditions~\cite{ntire26realx3d},
image denoising~\cite{ntire26denoising},
blind computational aberration correction~\cite{ntire26aberration},
event-based image deblurring~\cite{ntire26eventblurr},
efficient burst HDR and restoration~\cite{ntire26bursthdr},
low-light enhancement: `twilight cowboy'~\cite{ntire26twilight},
and efficient low light image enhancement~\cite{ntire26effllie}.

\section{NTIRE 2026 Remote Sensing Infrared Image Super-Resolution ($\times$4)}
The NTIRE 2026 Remote Sensing Infrared Image Super-Resolution Challenge, which is one of the associated challenges of NTIRE 2026, has two primary objectives. Firstly, it intends to offer a thorough overview of the latest advancements and emerging tendencies within the field of remote sensing infrared image super-resolution (SR). Secondly, it functions as a platform that enables both academic researchers and industrial professionals to converge and investigate possible collaborative opportunities for the practical application of infrared SR in remote sensing scenarios. The following part delves into the specific aspects.  

\vspace{-2.mm}
\subsection{Dataset}
\vspace{-1mm}
The challenge uses the official custom \textbf{InfraredSR} dataset (gratefully acknowledged to Spark Transmission (Beijing) Co., Ltd. for providing the data support). The low-resolution (LR)-high-resolution (HR) image pairs are constructed with the high-quality (HQ) infrared images via bicubic interpolation with a $\times$4 downsampling factor~\cite{li2026satvideodataset}.

The dataset is divided into training set (train), validation set (val) and test set (test), with the detailed image quantity and resolution distribution of each subset as follows:
320$\times$256: 625 images, 120$\times$120: 281 images, 64$\times$64: 99 images, 256$\times$256: 9 images, 160$\times$128: 5 images.

\noindent\textbf{Validation Set}
The validation set contains a total of 100 images, covering 5 resolution types:
320$\times$256: 66 images, 120$\times$120: 20 images, 64$\times$64: 12 images, 256$\times$256: 1 image, 160$\times$128: 1 image.

\noindent\textbf{Test Set}
The test set contains a total of 222 images, covering 3 resolution types:
1280$\times$1024: 188 images, 480$\times$480: 22 images, 256$\times$256: 12 images.

\subsection{Competition}
\label{sec:track}
This year's competition focuses on remote sensing infrared image super-resolution, with a unified evaluation metric for the competition.

\noindent\textbf{Infrared SR}.
All teams are ranked according to the comprehensive IQA score calculated from the enhanced HR images compared to the GT HR images of the InfraredSR testing dataset. The official IQA score calculation formula is defined as:
\begin{equation}
\begin{aligned}
        \text{Score} = \text{PSNR} + 20 \times \text{SSIM}.
\end{aligned}
\end{equation}

\noindent\textbf{Challenge Phases.}
\textit{(1) Development and Validation Phase:} During this phase, participants have access to the full InfraredSR training set and validation set. Participants train their models on the training set and submit the restored HR results on the validation set to the Codalab server for evaluation. The server will provide real-time ranking based on the official IQA score.

\textit{(2) Testing Phase:} In the final phase, participants receive the full InfraredSR test set (LR images only, with corresponding HR ground truth images kept confidential). They are required to upload their SR outputs (HR images) to the Codalab server and submit their code and a detailed report to the organizers via email. After the challenge ends, the organizers will validate the code and send the final evaluation results to the participants.

\noindent\textbf{Evaluation Protocol.} 
The official evaluation protocol and metric calculation scripts are available in the NTIRE 2026 official GitHub repository: \url{https://github.com/Kai-Liu001/NTIRE2026_infraredSR.git}, and the repository also includes the source code and pre-trained models. 
Key evaluation rules are as follows:
\begin{itemize}
    \item All metric calculations (PSNR and SSIM) are performed on the \textbf{single channel} of the infrared images (consistent with the characteristic of infrared image data).
    \item The final results shall be based on the official test results computed by the organizers. Code submitted by participants is used for reproduction and verification, with small discrepancies in precision being considered acceptable.
\end{itemize}

\begin{table*}[t]
    \centering
    \small
    \begin{adjustbox}{width=\linewidth}
    \setlength{\tabcolsep}{0.5cm}
    \renewcommand{\arraystretch}{0.95}

    \begin{tabular}{l|c|l|c|ccc}
        \toprule
        \textbf{Team Name} & \textbf{Rank} & \textbf{Leader} & \textbf{Email} & \textbf{Test PSNR} & \textbf{Test SSIM} & \textbf{Total Score} \\
        \midrule

        WHU-VIP        & 1  & Ce Wang           & wangcce6@gmail.com              & 35.9643 & 0.9236 & 54.4361 \\
        XJRes          & 2  & Li Pang           & pp2373886592@gmail.com         & 35.8713 & 0.9210 & 54.2920 \\
        FengFans       & 3  & Wei Zhou          & weichow@u.nus.edu              & 35.8211 & 0.9210 & 54.2416 \\
        I2WM\&JNU      & 4  & Weijun Yuan       & yweijun@stu2022.jnu.edu.cn     & 35.8263 & 0.9203 & 54.2319 \\
        davinci        & 5  & davinci           & davinci7571@gmail.com          & 35.8202 & 0.9204 & 54.2288 \\
        Earth4D        & 6  & Tongyao Mu        & muty.woodnr@gmail.com          & 35.7912 & 0.9202 & 54.1956 \\
        SUATSR         & 7  & Junye Chen        & suat24000219@stu.suat-sz.edu.cn & 35.7498 & 0.9198 & 54.1460 \\
        SiGMoid        & 8  & Kosuke Shigematsu & k-shigematsu@oita-ct.ac.jp     & 35.7001 & 0.9189 & 54.0784 \\
        hit\_zzy       & 9  & Zengyuan Zuo      & 3565741165@qq.com              & 35.6648 & 0.9185 & 54.0338 \\
        KLE Tech-CEVI  & 10 & Saiprasad Meesiyawar & saiprasad@cevi.co.in        & 35.6604 & 0.9184 & 54.0287 \\
        CASWiT\_SR\_IR & 11 & Antoine Carreaud  & antoine.carreaud@epfl.ch       & 35.6529 & 0.9182 & 54.0162 \\
        NTR            & 12 & Jiachen Tu        & jtu9@illinois.edu              & 35.6544 & 0.9180 & 54.0148 \\
        NUDT\_DeepIter & 13 & Xinhong Lan       & 2304946634@qq.com              & 35.5492 & 0.9172 & 53.8936 \\

        \bottomrule
    
    \end{tabular}
    \end{adjustbox}
    \caption{\textbf{Results of the NTIRE 2026 Infrared Image Super-Resolution Challenge (x4).}
    Teams are sorted by the official single-track rank. Leader and contact email are provided for each team.}

    \label{tab:main_results}
\end{table*}
\section{Challenge Results}
The challenge adopts a single-track evaluation protocol focusing on restoration quality. The results and rankings of all participating teams are summarized in Tab.~\ref{tab:main_results}.

\noindent\textbf{Overall Performance.} 
The WHU-VIP team achieves the best performance with a PSNR of 35.96 dB and SSIM of 0.9236, followed by XJRes and FengFans. 
The top-5 teams exhibit highly competitive results, with marginal differences within 0.15 dB, indicating that the leading approaches have converged to a similar level of reconstruction accuracy.
The SSIM values are consistently high, suggesting structural fidelity is well preserved. 
Moreover, the total score follows a similar trend with PSNR,  confirming restoration quality dominates the ranking criterion.

Further details on the evaluation protocol are provided in Sec.~\ref{sec:track}. 
The team order in Sec.~\ref{sec:teams} follows the same orders as presented in Tab.~\ref{tab:main_results}, where the top teams and their method are highlighted. 
Due to space limitations, the remaining teams are included in the supplementary material.

\subsection{Architectures and main ideas}
Throughout the challenge, participants proposed a variety of techniques tailored to the characteristics of infrared imagery. Here, we summarize several key design paradigms observed in the top-performing methods.

\begin{enumerate}

    \item \textbf{Transformer-based architectures for capturing global thermal structures.} 
    Transformer-based models remain dominant due to strong capability in modeling long-range dependencies. 
    In the context of infrared SR, these architectures are particularly effective for capturing global thermal distributions and structural patterns. 
    Many teams adopted pre-trained Transformer backbones such as HAT~\cite{chen2023activating}, SwinIR~\cite{liang2021swinir} and fine-tuned them to better adapt to the characteristics of infrared images.

    \vspace{2.mm}
    \item \textbf{State-space models (Mamba) for efficient long-range dependency modeling.} 
    To further enhance global context modeling with reduced computational overhead, several teams explored Mamba-based architectures. 
    These models are well-suited for infrared SR, where spatial correlations tend to be smoother and more globally distributed. 
    By integrating Mamba modules into Transformer frameworks, participants improved the modeling of large-scale thermal structures and achieved better reconstruction consistency.

    \vspace{2.mm}
    \item \textbf{Hybrid architectures and model fusion for balancing structure and detail.} 
    Given the inherent trade-off between structural fidelity and detail enhancement in infrared images, many teams employed hybrid architectures combining Transformers and CNNs. 
    Transformer branches focus on global structure recovery, while CNN branches emphasize local detail refinement and noise suppression. 
    Ensemble and fusion strategies further improved overall reconstruction performance.

    \vspace{2.mm}
    \item \textbf{Frequency-aware learning for enhancing weak high-frequency signals.} 
    Unlike visible images, infrared images often lack strong high-frequency components. 
    To address this, participants incorporated frequency-domain supervision to explicitly enhance subtle structural details and edges. 
    These techniques help mitigate over-smoothing and improve perceptual sharpness without introducing artifacts.

    \vspace{2.mm}
    \item \textbf{Progressive and curriculum-based training strategies.} 
    Multi-stage training strategies were widely adopted to stabilize optimization and improve performance. 
    Typically, models are first trained on smaller patches or easier settings and then progressively refined using more challenging data distributions. 
    This approach is particularly beneficial for infrared SR, where data diversity is limited and model generalization is critical.

\end{enumerate}

\subsection{Participants}
This year, the image SR challenge attracted 115 registered participants, with 13 teams submitting valid entries. These submissions have set a new benchmark for the state-of-the-art in infrared image super-resolution ($\times$4).

\subsection{Fairness}
To ensure a fair and reliable comparison across all participating methods, the following rules are enforced throughout the challenge.
\textbf{(1)} The use of ground-truth high-resolution (HR) images from the validation and test sets for training is strictly prohibited. Only the provided training data can be used for supervised learning, while the test HR images remain confidential.
\textbf{(2)} Participants are allowed to utilize additional external datasets for training, provided that no overlap exists with the InfraredSR validation and test sets. Any form of data leakage from the test set is strictly forbidden.
\textbf{(3)} All evaluation metrics are computed on the single-channel infrared images using the official evaluation protocol. Participants must ensure that their methods follow the same preprocessing and evaluation settings.
\textbf{(4)} Final rankings are determined based on the official results computed by the organizers on the hidden test set. Submitted codes are used for reproducibility verification.

\subsection{Conclusions}
The insights gained from analyzing the results of the NTIRE 2026 Infrared Image Super-Resolution Challenge are summarized as follows:
\begin{enumerate}

    \item \textbf{Hybrid architectures effectively balance global structure and local detail reconstruction.}
    The combination of Transformer-based models and CNNs remains a dominant design paradigm.
    Transformers capture global thermal structures and long-range dependencies, while CNNs enhance local detail refinement and suppress noise, leading to consistently strong restoration performance on infrared images.

    \item \textbf{Efficient long-range modeling is critical for infrared imagery.}
    State-space models such as Mamba have shown promising capability in modeling smooth and globally distributed spatial correlations. 
    These approaches provide a computationally efficient alternative to self-attention, particularly suitable for infrared data with limited high-frequency content.

    \item \textbf{Training strategies play a key role in improving robustness and generalization.}
    Multi-stage and curriculum-based training schemes, such as progressive patch size adjustment, are widely adopted to stabilize optimization. 
    These strategies are especially beneficial for infrared SR, where data diversity is limited and overfitting can easily occur.

    \item \textbf{Structure-aware and frequency-aware learning improves detail reconstruction.}
    Given the inherently weak high-frequency signals in infrared images, incorporating frequency-domain supervision and structure-aware objectives helps enhance edge sharpness and mitigate over-smoothing effects.

\end{enumerate}

\section*{Acknowledgments}
This work was partially supported by the Humboldt Foundation. We thank the NTIRE 2026 sponsors: OPPO, Kuaishou, and the University of Wurzburg (Computer Vision Lab).

\section{Challenge Methods and Teams}
\label{sec:teams}

\subsection{WHU-VIP}
\begin{figure*}[htb]
    \centering
    \includegraphics[width=\textwidth]{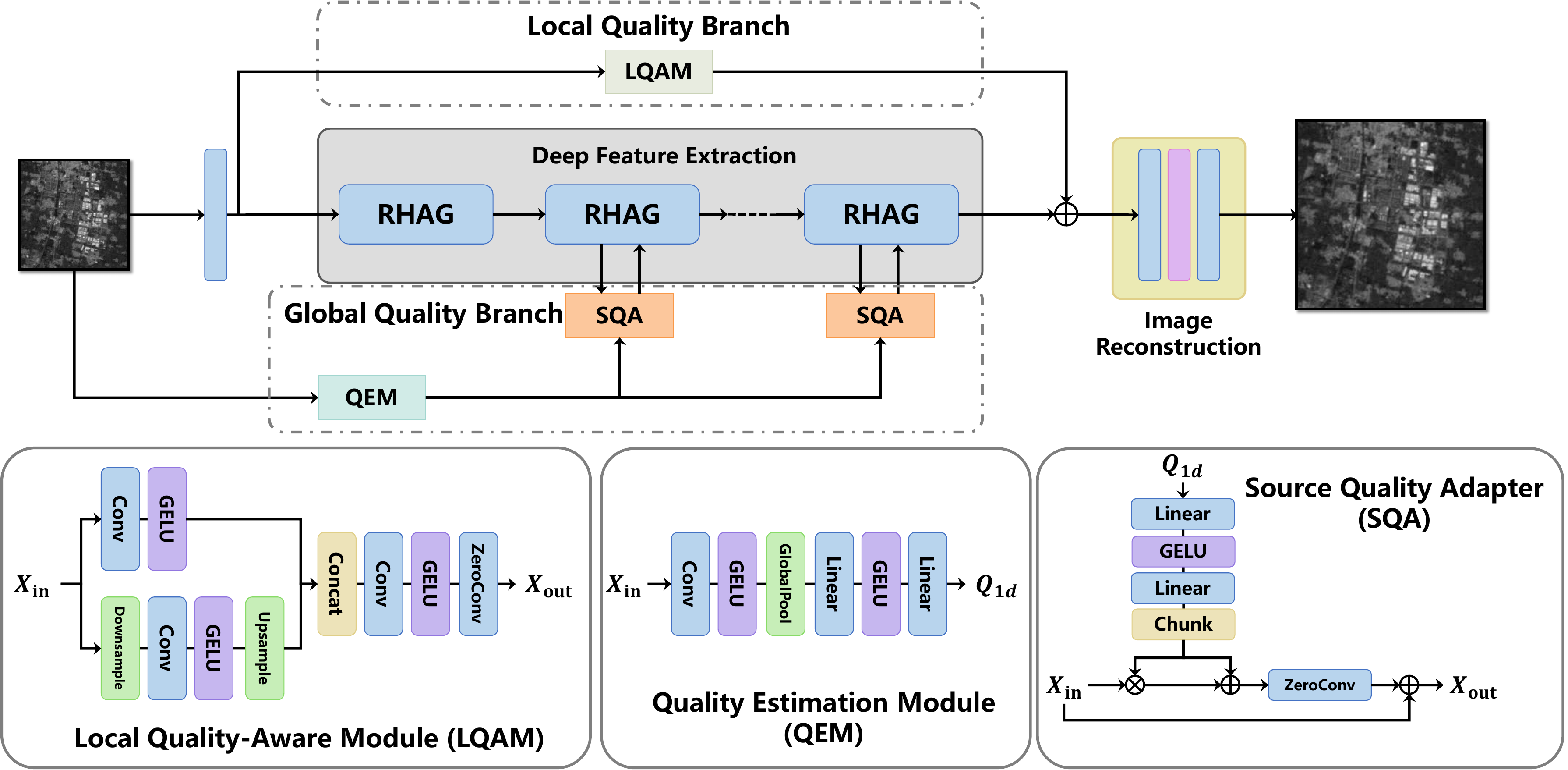}
    \caption{Architecture of the proposed Quality-Aware Hybrid Attention Transformer (QAHAT).}
    \label{fig:team11}
\end{figure*}
\noindent\textbf{Description.}
Their approach is built upon the Hybrid Attention Transformer (HAT) \cite{chen2023activating}. In the challenge dataset, the low-resolution (LR) images are generated from high-resolution (HR) images using hand-crafted interpolation. However, the HR images themselves exhibit significant quality variations. Specifically, some HR images correspond to true high-resolution satellite observations, while others originate from lower-resolution sources or contain noticeable degradations such as sensor noise or JPEG compression artifacts. As a result, the HR supervision is heterogeneous, and treating all training samples equally may introduce inconsistent learning signals.

To address this issue, they propose a Quality-Aware HAT (QAHAT) that explicitly models the quality variations of HR images which is shown in \cref{fig:team11}. Their framework introduces two complementary components: a Global Quality Branch and a Local Quality Branch. The global branch estimates the overall degradation characteristics of the input, while the local branch captures spatially varying artifacts. By incorporating both global and local quality cues, the proposed method enables the backbone network to adapt its feature processing to different degradation conditions.

They introduce a lightweight Quality Estimation Module (QEM) to estimate the overall degradation of the input image. The QEM consists of several strided convolutional layers followed by global pooling and fully connected layers, producing a compact quality descriptor that summarizes the degradation characteristics of the input image. The estimated quality vector is injected into the backbone through multiple Source Quality Adapters (SQA) inserted in the deep feature extraction stages of HAT. Each SQA modulates intermediate feature maps using channel-wise scaling and bias parameters derived from the quality representation, allowing the network to dynamically adjust feature responses according to the predicted degradation level.

In addition to global quality variations, infrared remote sensing images may contain spatially localized artifacts. To model these local degradations, they introduce an Local Quality-Aware Module (LQAM). The LQAM extracts artifact-aware features using a downsampling pathway that provides a larger receptive field for capturing contextual degradation patterns. The resulting artifact representation is fused with the backbone features to guide the reconstruction process in regions affected by degradations. Similar to the global branch, the fused features are integrated into the HAT backbone through a residual connection to ensure stable training.

By jointly modeling global degradation characteristics and local artifact patterns, the proposed QAHAT improves the robustness of the process and produces more reliable reconstructions for infrared remote sensing images.

\noindent\textbf{Implementation Details.}
{\textit{Datasets.}}In addition to the training data provided by the challenge, they incorporate the publicly available thermal infrared remote sensing dataset SatVideoIRSDT \cite{li2025probing} as supplementary training data. Since SatVideoIRSDT is a video dataset, they randomly sample one frame from each video sequence to construct an auxiliary thermal infrared image dataset. The sampled frames are then combined with the official competition training data to form the final training set.

{\textit{Trainning strategy.}} The proposed model is trained on image patches of size $256 \times 256$. They adopt the Adam optimizer with an initial learning rate of $5\times10^{-5}$ and momentum parameters $\beta_1 = 0.9$ and $\beta_2 = 0.99$. The learning rate is updated using a MultiStepLR scheduler, where the learning rate is reduced by a factor of 0.5 at 20k and 40k iterations. The model is trained for 200k iterations in total. An exponential moving average (EMA) with a decay factor of 0.999 is maintained to stabilize the training process. Following the evaluation protocol of the challenge, the training objective is designed to align with the official metric. The loss function is defined as $\mathcal{L} = -\text{PSNR}(I_{\text{sr}}, I_{\text{gt}}) - 20 \cdot \text{SSIM}(I_{\text{sr}}, I_{\text{gt}})$. The backbone network is initialized using pretrained weights of the HAT-L model, while the newly introduced modules are randomly initialized and jointly optimized during training.

\subsection{XJRes}

\noindent\textbf{Description.}
The proposed method combines two complementary super-resolution models, namely the Progressive Focused Transformer (PFT \cite{Long_2025_CVPR}) and the Hybrid Attention Transformer (HAT \cite{chen2023activating}). Given a low-resolution input image $I_{LR}$, the image is processed by four parallel branches. The first two branches adopt the PFT \cite{Long_2025_CVPR} architecture, while the last two branches employ the HAT \cite{chen2023activating} architecture to generate super-resolved results. The outputs from these four branches are then fused through a weighted averaging strategy to obtain the final high-resolution image. This design enables the framework to simultaneously exploit the progressive attention focusing capability of PFT \cite{Long_2025_CVPR} and the hybrid attention representation ability of HAT \cite{chen2023activating}, thereby enhancing the reconstruction performance.

\begin{figure*}[ht]
    \centering
    \includegraphics[width=.7\textwidth]{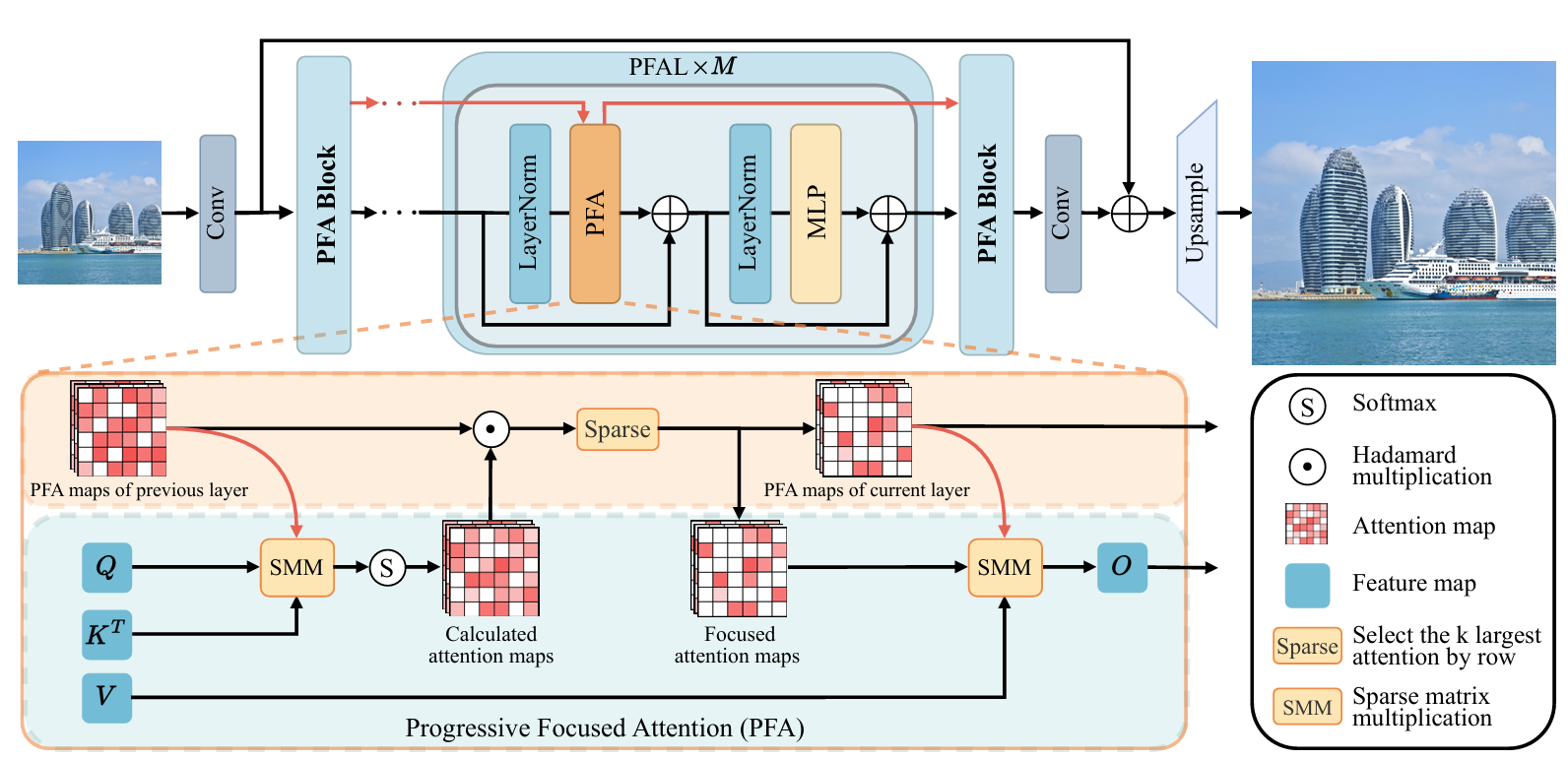}
    \caption{Overall architecture of the Progressive Focused Transformer (PFT).}
    \label{fig:PFT}
\end{figure*}

The first two branches adopt the Progressive Focused Transformer (PFT \cite{Long_2025_CVPR}) architecture for image super-resolution. Note that these two PFT \cite{Long_2025_CVPR} branches share the same network architecture but are trained independently with different objective functions to capture complementary features. As illustrated in Fig.~\ref{fig:PFT}, the network progressively extracts image features and then reconstructs the high-resolution output.

\begin{itemize}

\item \textbf{Feature Encoding:}  
Given the low-resolution input image $I_{LR}$, a convolutional layer is first applied to transform the input into a high-dimensional feature representation $F_{0}$. This stage extracts basic structural and texture information for subsequent feature learning.

\item \textbf{Progressive Attention Representation Learning:}  
The core of this branch consists of multiple Progressive Focused Attention (PFA) blocks. Each block contains several Progressive Focused Attention Layers (PFAL), where attention maps are propagated across layers to progressively refine the focus on important tokens. Through attention inheritance and sparse selection, the PFA mechanism enhances highly relevant features while suppressing irrelevant tokens, enabling efficient feature aggregation.

\item \textbf{Image Reconstruction:}  
After deep feature extraction, the refined features are passed through convolutional layers and an upsampling module to generate a super-resolved image $I_{SR}^{PFT}$.

\end{itemize}

The remaining two branches are based on the Hybrid Attention Transformer (HAT \cite{chen2023activating}). Similarly, these two HAT branches share the same network architecture but are trained independently with different objective functions. The model integrates both channel attention and window-based self-attention to improve feature representation.

\begin{itemize}

\item \textbf{Shallow Feature Extraction:}  
The input image $I_{LR}$ is first processed by a $3\times3$ convolutional layer to obtain shallow features $F_{0}$, mapping the image into a higher-dimensional feature space.

\item \textbf{Deep Feature Extraction:}  
This stage contains several Residual Hybrid Attention Groups (RHAG), each consisting of multiple Hybrid Attention Blocks (HAB) and an Overlapping Cross-Attention Block (OCAB). The HAB combines window-based self-attention and channel attention to capture both local textures and global dependencies, while the OCAB enhances cross-window interactions for more effective feature aggregation.

\item \textbf{Image Reconstruction:}  
Finally, a convolution layer followed by a pixel-shuffle \cite{shi2016real} operation upsamples the deep features to produce the super-resolved output $I_{SR}^{HAT}$.

\end{itemize}

After obtaining the outputs from the four branches, the final super-resolved image is generated by a weighted fusion strategy combined with a self-ensemble technique during inference, which aggregates predictions from geometrically transformed inputs~\cite{timofte2016seven}. Let $I_{SR}^{(i)}$ denote the super-resolved output of the $i$-th branch. The final result is computed as the weighted average of the four outputs:

\begin{equation}
I_{SR} = 0.25\, I_{SR}^{(1)} + 0.25\, I_{SR}^{(2)} + 0.25\, I_{SR}^{(3)} + 0.25\, I_{SR}^{(4)}.
\end{equation}
where the first two branches correspond to the PFT model and the remaining two branches correspond to the HAT model. Since all weights are equal, the fusion operation is equivalent to averaging the outputs of the four branches.

\noindent\textbf{Implementation Details.}
\begin{itemize}[leftmargin=*,noitemsep]
  \item \textbf{Data:} 1,019 official HR/LR pairs, no extra data
  \item \textbf{Loss:} PFT branches: RBSFormer \cite{li2024rbsformer} \& $L_1$; HAT branches: RBSFormer \cite{li2024rbsformer} \& CharbonnierL1 + SSIM
  \item \textbf{Training:} lr=$2 \times 10^{-4}$ (Cosine Annealing), 200ep, bs=1, FP32
  \item \textbf{Optim:} Adam \cite{kingma2014adam}
  \item \textbf{HW:} RTX 4090 GPUs (PyTorch)
\end{itemize}

\subsection{FengFans}
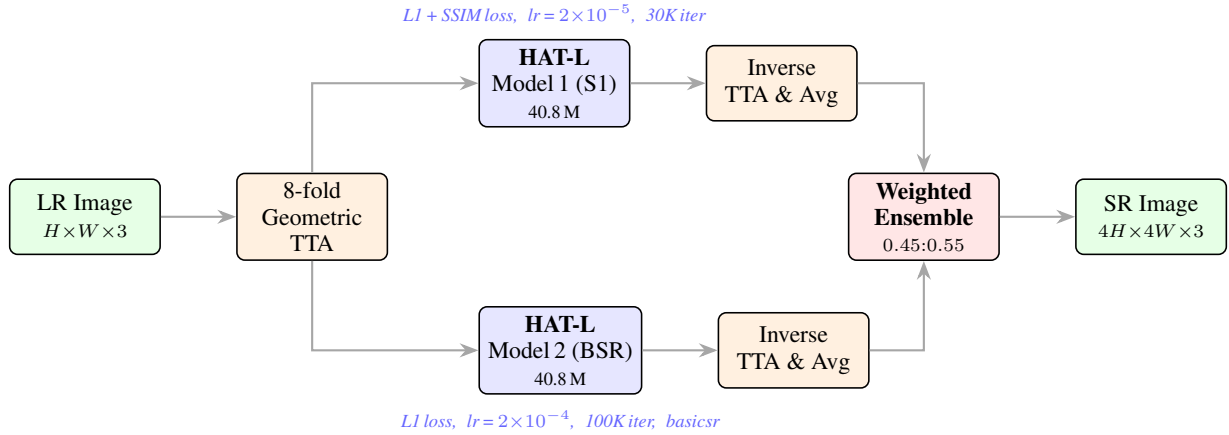
\begin{figure*}[t]
\centering
\begin{tikzpicture}[
    node distance=0.55cm and 0.7cm,
    block/.style={rectangle, draw, rounded corners=3pt,
                  minimum height=1cm, minimum width=2cm,
                  align=center, font=\small},
    modelblock/.style={block, fill=blue!10, line width=0.6pt},
    ttablock/.style={block, fill=orange!12, line width=0.6pt},
    ioblock/.style={block, fill=green!10, line width=0.6pt},
    ensblock/.style={block, fill=red!10, line width=0.6pt},
    arrow/.style={-{Stealth[length=2.5mm]}, thick, color=gray!70},
    label/.style={font=\scriptsize\itshape, text=blue!60},
]
\node[ioblock] (lr) {LR Image\\[-1pt]{\scriptsize$H{\times}W{\times}3$}};
\node[ttablock, right=1cm of lr] (tta) {8-fold\\[-1pt]Geometric\\[-1pt]TTA};
\node[modelblock, above right=0.6cm and 1.2cm of tta] (m1) {\textbf{HAT-L}\\[-1pt]Model\,1 (S1)\\[-1pt]{\scriptsize 40.8\,M}};
\node[modelblock, below right=0.6cm and 1.2cm of tta] (m2) {\textbf{HAT-L}\\[-1pt]Model\,2 (BSR)\\[-1pt]{\scriptsize 40.8\,M}};
\node[ttablock, right=1cm of m1] (inv1) {Inverse\\[-1pt]TTA \& Avg};
\node[ttablock, right=1cm of m2] (inv2) {Inverse\\[-1pt]TTA \& Avg};
\node[ensblock, right=0.8cm of $(inv1)!0.5!(inv2)$] (ens) {\textbf{Weighted}\\[-1pt]\textbf{Ensemble}\\[-1pt]{\scriptsize$0.45{:}0.55$}};
\node[ioblock, right=1cm of ens] (sr) {SR Image\\[-1pt]{\scriptsize$4H{\times}4W{\times}3$}};
\draw[arrow] (lr) -- (tta);
\draw[arrow] (tta) |- (m1);
\draw[arrow] (tta) |- (m2);
\draw[arrow] (m1) -- (inv1);
\draw[arrow] (m2) -- (inv2);
\draw[arrow] (inv1) -| (ens);
\draw[arrow] (inv2) -| (ens);
\draw[arrow] (ens) -- (sr);
\node[label, above=0.08cm of m1] {L1\,+\,SSIM\,loss,\; lr\,=\,$2{\times}10^{-5}$,\; 30K\,iter};
\node[label, below=0.08cm of m2] {L1\,loss,\; lr\,=\,$2{\times}10^{-4}$,\; 100K\,iter,\; basicsr};
\end{tikzpicture}
\caption{\textbf{Inference pipeline of their solution.} A low-resolution infrared image is augmented with 8 geometric transforms (4~rotations\,$\times$\,2~flips). Each augmented copy is independently super-resolved by two HAT-L models trained with different strategies. The per-model outputs are inverse-transformed and averaged, then the two model predictions are fused with learned weights (0.45\,:\,0.55) to produce the final $\times$4 SR output.}
\label{fig:pipeline}
\end{figure*}

\noindent\textbf{Description.}
Their solution for the NTIRE 2026 Remote Sensing Infrared Image Super-Resolution ($\times$4) challenge is based on a \emph{weighted ensemble} of two HAT-L (Hybrid Attention Transformer -- Large)~\cite{chen2023activating} models combined with 8-fold geometric Test-Time Augmentation~(TTA)~\cite{timofte2016seven}. Although both models share the identical HAT-L architecture (40.8\,M parameters each), they are trained with complementary optimization strategies: one emphasizes \emph{structural fidelity} via an SSIM-aware loss, while the other targets \emph{pixel-level accuracy} with a pure L1 loss and substantially longer training. The resulting predictions are diverse enough that their weighted average consistently outperforms either individual model. Figure~\ref{fig:pipeline} depicts the full inference pipeline.

They adopt HAT-L~\cite{chen2025hat}, a state-of-the-art hybrid attention transformer for image super-resolution. HAT augments standard window-based self-attention with two complementary mechanisms:

\begin{enumerate}[leftmargin=*,nosep]
\item \textbf{Overlapping cross-attention (OCA):} adjacent windows share key--value regions with a configurable overlap ratio, enabling information exchange across window boundaries without the quadratic cost of global attention.
\item \textbf{Channel attention block (CAB):} a squeeze-and-excitation style module integrated within each transformer block recalibrates channel-wise feature responses, improving representational capacity.
\end{enumerate}

\noindent The ``Large'' variant stacks 12 Residual Hybrid Attention Groups (RHAGs), each containing 6 Hybrid Attention Blocks (HABs). Table~\ref{tab:arch} lists the full configuration.

\begin{table}[t]
\centering
\small
\caption{HAT-L network configuration.}
\label{tab:arch}
\setlength{\tabcolsep}{4pt}
\begin{tabular}{@{}lc@{}}
\toprule
\textbf{Hyperparameter} & \textbf{Value} \\
\midrule
Input / output channels & 3 \\
Embedding dimension & 180 \\
\# Residual Groups (RHAG) & 12 \\
\# HABs per RHAG & 6 \\
\# Attention heads & 6 \\
Window size & $16{\times}16$ \\
Overlap ratio (OCA) & 0.5 \\
MLP expansion ratio & 2 \\
Channel compress ratio & 3 \\
Squeeze factor (CAB) & 30 \\
Conv scale & 0.01 \\
Upsampler & PixelShuffle ($\times$4) \\
Image value range & $[0,\,1]$ \\
\midrule
\# Parameters & 40.8\,M \\
\bottomrule
\end{tabular}
\end{table}

Although the competition images are single-channel infrared data, they represent them as 3-channel RGB by replicating the gray channel. This allows us to directly leverage pretrained HAT-L weights without any architectural modification. The official evaluation converts both the SR output and the ground truth to a single luminance channel via $Y = 0.299R + 0.587G + 0.114B$ before computing PSNR and SSIM.

Pretraining: Both models are initialized from the official HAT-L checkpoint that was first pretrained on ImageNet for image classification and then trained for $\times$4 super-resolution on the DF2K dataset (DIV2K\,+\,Flickr2K, $\approx$3{,}450 high-resolution images). This two-stage pretraining endows the backbone with strong low-level feature representations for natural images.

They fine-tune on the competition's 1{,}019 paired HR--LR infrared training images. Two models are trained with \emph{deliberately different} hyperparameters to maximize ensemble diversity. Table~\ref{tab:training} compares the two configurations.

\begin{table}[t]
\centering
\small
\caption{Training configurations of the two models.}
\label{tab:training}
\setlength{\tabcolsep}{3pt}
\begin{tabular}{@{}lcc@{}}
\toprule
 & \textbf{Model\,1 (S1)} & \textbf{Model\,2 (BSR)} \\
\midrule
Loss & L1 + 0.5\,$\mathcal{L}_{\text{SSIM}}$ & L1 \\
Optimizer & AdamW & Adam \\
Initial LR & $2{\times}10^{-5}$ & $2{\times}10^{-4}$ \\
LR schedule & Cosine & MultiStep \\
 & & {\scriptsize(50/80/90/95K)} \\
Warmup & 1{,}000 iter & None \\
Total iterations & 30{,}000 & 100{,}000 \\
Patch size & $256{\times}256$ & $256{\times}256$ \\
Batch size & 4 & 4 \\
EMA decay & 0.999 & 0.999 \\
Weight decay & 0 & 0 \\
Gradient clip & 1.0 & -- \\
Random seed & 42 & 42 \\
Framework & Custom & basicsr \\
\bottomrule
\end{tabular}
\end{table}

The competition scoring formula $\text{Score} = \text{PSNR} + 20 \times \text{SSIM}$ amplifies SSIM by a factor of 20 relative to PSNR. Concretely, an SSIM improvement of 0.01 contributes the same as a 0.2\,dB PSNR gain. They therefore incorporate a differentiable SSIM loss term ($\mathcal{L}_{\text{SSIM}} = 1 - \text{SSIM}$) alongside L1 to encourage the model to preserve structural patterns---particularly important for infrared imagery where thermal edge boundaries and fine textures carry critical information. The low learning rate ($2{\times}10^{-5}$) with cosine annealing prevents catastrophic forgetting of the pretrained features.

A 10$\times$ higher initial learning rate ($2{\times}10^{-4}$) combined with a stepwise decay schedule enables more aggressive adaptation to the infrared domain. The substantially longer training (100K iterations, \emph{vs.}\ 30K for S1) allows the model to thoroughly learn infrared-specific data statistics. Using pure L1 loss produces different gradient dynamics than the SSIM-augmented loss, yielding a model with complementary error patterns that benefits the ensemble.

Data Augmentation: Both models share the same augmentation pipeline:
\begin{itemize}[leftmargin=*,nosep]
\item Random horizontal flip ($p = 0.5$)
\item Random vertical flip ($p = 0.5$)
\item Random rotation from $\{90^{\circ}, 180^{\circ}, 270^{\circ}\}$
\end{itemize}
They intentionally omit color-space augmentations (jitter, channel shuffle) and advanced mixing strategies (MixUp, CutMix), because the single-channel infrared images have fundamentally different intensity statistics from natural RGB photographs.

Test-Time Augmentation (TTA)~\cite{timofte2016seven}: For each input LR image, they apply all 8 geometric transforms from the dihedral group $D_4$ (4 rotations $\times$ 2 flips). Each transformed copy is super-resolved, inverse-transformed back to the original orientation, and the 8 outputs are averaged:
\begin{equation}
\hat{I}_{m} = \frac{1}{8}\sum_{k=1}^{8} T_k^{-1}\!\bigl(f_m\!\bigl(T_k(I_{\text{LR}})\bigr)\bigr),
\end{equation}
where $f_m$ is model $m$ and $T_k$ is the $k$-th transform. TTA is their single largest source of test-time improvement (+2.54 points on the test set).

Weighted Model Ensemble: The final SR output fuses the two per-model predictions with optimized weights:
\begin{equation}
\hat{I}_{\text{SR}} = \alpha\,\hat{I}_{1} + (1-\alpha)\,\hat{I}_{2},\quad \alpha = 0.45.
\end{equation}

The weight $\alpha$ was selected through a systematic grid search over $[0.30, 0.60]$, evaluating each configuration on the test leaderboard.  Table~\ref{tab:weights} reports the sensitivity.

\begin{table}[t]
\centering
\small
\caption{Ensemble weight sensitivity ($\alpha$ = S1 weight). Score~=~PSNR + 20\,$\times$\,SSIM on the test set.}
\label{tab:weights}
\setlength{\tabcolsep}{5pt}
\begin{tabular}{@{}cccc@{}}
\toprule
$\alpha$ (S1) & PSNR & SSIM & Score \\
\midrule
0.40 & 35.82 & 0.9205 & 54.228 \\
0.42 & 35.81 & 0.9207 & 54.228 \\
0.43 & 35.81 & 0.9207 & 54.228 \\
\textbf{0.45} & \textbf{35.81} & \textbf{0.9207} & \textbf{54.229} \\
0.50 & 35.81 & 0.9208 & 54.222 \\
0.55 & 35.80 & 0.9211 & 54.219 \\
0.60 & 35.78 & 0.9213 & 54.208 \\
\bottomrule
\end{tabular}

\end{table}
\noindent The optimal weight slightly favors BSR100K ($1-\alpha = 0.55$), consistent with its stronger pixel-level accuracy from longer training. Notably, the score is robust within $\alpha \in [0.40, 0.50]$, varying by only 0.007 across this range.

\noindent\textbf{Implementation Details.}
\begin{itemize}[leftmargin=*,nosep]
\item \textbf{No mixed precision:} They discovered that HAT-L produces NaN outputs when using fp16 autocast (likely due to overflow in the attention softmax with 16$\times$16 windows). All training and inference therefore use fp32.
\item \textbf{Reflect padding:} Input images are padded to the nearest multiple of the window size (16) using reflection padding before inference, then the output is cropped to the exact target resolution ($4H \times 4W$).
\item \textbf{No tiling:} All test images have LR dimensions $\leq 320 \times 256$, which comfortably fit in GPU memory without tile-based processing.
\end{itemize}

\vspace{-2.mm}
\section{Methods of the Remaining Teams}
\vspace{-2.mm}
All the teams presented innovative ideas and thorough experiments for the competition. However, due to space limitations, a more in-depth discussion can be found in Sec.~A of the supplementary materials, which contains detailed descriptions of the methods and implementation details for the remaining teams that participated in the challenge. While these teams have not been discussed in the main report, their approaches are still highlighted, offering insight into their unique strategies and technical implementations.

\section*{Acknowledgments}
This work is supported by the National Natural Science Foundation of China (62501386, 625B2116, 625B1025), CCF-Tencent Rhino-Bird Open Research Fund.
This work is also sponsored by Al Hundred Schools Program and is carried out using the Ascend AI technology stack.
This work was partially supported by the Humboldt Foundation. We thank the NTIRE 2026 sponsors: OPPO, Kuaishou, and the University of Wurzburg (Computer Vision Lab).

{\small
\bibliographystyle{ieeenat_fullname}
\bibliography{main}

@String(CVPR= {IEEE Conf. Comput. Vis. Pattern Recog.})

@String(ICCV= {Int. Conf. Comput. Vis.})

@String(ECCV= {Eur. Conf. Comput. Vis.})

@String(CVPRW= {IEEE Conf. Comput. Vis. Pattern Recog. Worksh.})

@String(CVPR  = {CVPR})

@String(ICCV  = {ICCV})

@String(ECCV  = {ECCV})

@String(CVPRW= {CVPRW})

@inproceedings{dong2014learning,
  title={Learning a deep convolutional network for image super-resolution},
  author={Dong, Chao and Loy, Chen Change and He, Kaiming and Tang, Xiaoou},
  booktitle={ECCV},
  year={2014}
}

@inproceedings{shi2016real,
  title={Real-time single image and video super-resolution using an efficient sub-pixel convolutional neural network},
  author={Shi, Wenzhe and Caballero, Jose and Husz{\'a}r, Ferenc and Totz, Johannes and Aitken, Andrew P and Bishop, Rob and Rueckert, Daniel and Wang, Zehan},
  booktitle={CVPR},
  year={2016}
}

@inproceedings{ledig2017photo,
  title={Photo-realistic single image super-resolution using a generative adversarial network},
  author={Ledig, Christian and Theis, Lucas and Husz{\'a}r, Ferenc and Caballero, Jose and Cunningham, Andrew and Acosta, Alejandro and Aitken, Andrew and Tejani, Alykhan and Totz, Johannes and Wang, Zehan and Shi, Wenzhe},
  booktitle={CVPR},
  year={2017}
}

@inproceedings{goodfellow2014generative,
  title={Generative adversarial nets},
  author={Goodfellow, Ian and Pouget-Abadie, Jean and Mirza, Mehdi and Xu, Bing and Warde-Farley, David and Ozair, Sherjil and Courville, Aaron and Bengio, Yoshua},
  booktitle={NeurIPS},
  year={2014}
}

@inproceedings{zhang2018residual,
  title={Residual Dense Network for Image Super-Resolution},
  author={Zhang, Yulun and Tian, Yapeng and Kong, Yu and Zhong, Bineng and Fu, Yun},
  booktitle={CVPR},
  year={2018}
}

@InProceedings{chen2023activating,
    author    = {Chen, Xiangyu and Wang, Xintao and Zhou, Jiantao and Qiao, Yu and Dong, Chao},
    title     = {Activating More Pixels in Image Super-Resolution Transformer},
    booktitle = {CVPR},
    year      = {2023},
}

@inproceedings{liang2021swinir,
  title={Swinir: Image restoration using swin transformer},
  author={Liang, Jingyun and Cao, Jiezhang and Sun, Guolei and Zhang, Kai and Van Gool, Luc and Timofte, Radu},
  booktitle={ICCVW},
  year={2021}
}

@article{kingma2014adam,
  title={Adam: A method for stochastic optimization},
  author={Kingma, Diederik P and Ba, Jimmy},
  journal={arXiv preprint arXiv:1412.6980},
  year={2014}
}

@article{vaswani2017attention,
  title={Attention is all you need},
  author={Vaswani, Ashish and Shazeer, Noam and Parmar, Niki and Uszkoreit, Jakob and Jones, Llion and Gomez, Aidan N and Kaiser, {\L}ukasz and Polosukhin, Illia},
  journal={NeurIPS},
  year={2017}
}

@article{liu2024vmamba,
  title={Vmamba: Visual state space model},
  author={Liu, Yue and Tian, Yunjie and Zhao, Yuzhong and Yu, Hongtian and Xie, Lingxi and Wang, Yaowei and Ye, Qixiang and Liu, Yunfan},
  journal={arXiv preprint arXiv:2401.10166},
  year={2024}
}

@inproceedings{chen2023dual,
  title={Dual Aggregation Transformer for Image Super-Resolution},
  author={Chen, Zheng and Zhang, Yulun and Gu, Jinjin and Kong, Linghe and Yang, Xiaokang and Yu, Fisher},
  booktitle={ICCV},
  year={2023}
}

@inproceedings{kim2016accurate,
  title={Accurate image super-resolution using very deep convolutional networks},
  author={Kim, Jiwon and Kwon Lee, Jung and Mu Lee, Kyoung},
  booktitle={CVPR},
  year={2016}
}

@inproceedings{zhang2018image,
  title={Image super-resolution using very deep residual channel attention networks},
  author={Zhang, Yulun and Li, Kunpeng and Li, Kai and Wang, Lichen and Zhong, Bineng and Fu, Yun},
  booktitle={ECCV},
  year={2018}
}

@inproceedings{dai2019second,
  title={Second-order Attention Network for Single Image Super-Resolution},
  author={Dai, Tao and Cai, Jianrui and Zhang, Yongbing and Xia, Shu-Tao and Zhang, Lei},
  booktitle={CVPR},
  year={2019}
}

@inproceedings{liu2021swin,
  title={Swin transformer: Hierarchical vision transformer using shifted windows},
  author={Liu, Ze and Lin, Yutong and Cao, Yue and Hu, Han and Wei, Yixuan and Zhang, Zheng and Lin, Stephen and Guo, Baining},
  booktitle={ICCV},
  year={2021}
}

@article{gu2023mamba,
  title={Mamba: Linear-time sequence modeling with selective state spaces},
  author={Gu, Albert and Dao, Tri},
  journal={arXiv preprint arXiv:2312.00752},
  year={2023}
}

@inproceedings{guo2024mambair,
  title={{MambaIR}: {A} simple baseline for image restoration with state-space model},
  author={Guo, Hang and Li, Jinmin and Dai, Tao and Ouyang, Zhihao and Ren, Xudong and Xia, Shu-Tao},
  booktitle=ECCV,
  pages={222--241},
  year={2025},
  organization={Springer}
}

@inproceedings{ho2020denoising,
  title={Denoising diffusion probabilistic models},
  author={Ho, Jonathan and Jain, Ajay and Abbeel, Pieter},
  booktitle={NeurIPS},
  year={2020}
}

@article{xia2023diffir,
  title={DiffIR: Efficient Diffusion Model for Image Restoration},
  author={Xia, Bin and Zhang, Yulun and Wang, Shiyin and Wang, Yitong and Wu, Xinglong and Tian, Yapeng and Yang, Wenming and Van Gool, Luc},
  journal={arXiv preprint arXiv:2303.09472},
  year={2023}
}

@article{saharia2022image,
  title={Image super-resolution via iterative refinement},
  author={Saharia, Chitwan and Ho, Jonathan and Chan, William and Salimans, Tim and Fleet, David J and Norouzi, Mohammad},
  journal={TPAMI},
  year={2022},
}

@article{li2024distillation,
  title={Distillation-Free One-Step Diffusion for Real-World Image Super-Resolution},
  author={Li, Jianze and Cao, Jiezhang and Zou, Zichen and Su, Xiongfei and Yuan, Xin and Zhang, Yulun and Guo, Yong and Yang, Xiaokang},
  journal={arXiv preprint arXiv:2410.04224},
  year={2024}
}

@inproceedings{wu2023seesr,
  title={SeeSR: Towards Semantics-Aware Real-World Image Super-Resolution},
  author={Wu, Rongyuan and Yang, Tao and Sun, Lingchen and Zhang, Zhengqiang and Li, Shuai and Zhang, Lei},
  booktitle={CVPR},
  year={2024}
}

@article{li2026satvideodataset,
  title={Infrared video satellite aerial moving target detection dataset and its evaluation},
  author={Li, Ruojing and Li, Zhaoxu and Chen, Nuo and Guo, Gaowei and Dou, Zechao and Long, Zhengxing and Luo, Yihang and Zeng, Yaoyuan and Sheng, Weidong and Li, Boyang and others},
  doi={10.11834/jig.250536},
  journal={Journal of Image and Graphics},
  pages={1--15},
  year={2026}
}

@article{li2025probing,
  title={Probing deep into temporal profile makes the infrared small target detector much better},
  author={Li, Ruojing and An, Wei and Wang, Yingqian and Ying, Xinyi and Dai, Yimian and Wang, Longguang and Li, Miao and Guo, Yulan and Liu, Li},
  journal={arXiv preprint arXiv:2506.12766},
  year={2025}
}

@inproceedings{Long_2025_CVPR,
  title     = {Progressive Focused Transformer for Single Image Super-Resolution},
  author    = {Long, Wei and Zhou, Xingyu and Zhang, Leheng and Gu, Shuhang},
  booktitle = {Proceedings of the IEEE/CVF Conference on Computer Vision and Pattern Recognition (CVPR)},
  pages     = {2279--2288},
  year      = {2025}
}

@inproceedings{li2024rbsformer,
  title     = {{RBSFormer}: Ray Beam Search Transformer for RAW Image Super-Resolution},
  author    = {Li, B. and others},
  booktitle = {Proceedings of the IEEE/CVF Conference on Computer Vision and Pattern Recognition Workshops (CVPRW)},
  year      = {2024}
}

@article{chen2025hat,
  title={Hat: Hybrid attention transformer for image restoration},
  author={Chen, Xiangyu and Wang, Xintao and Zhang, Wenlong and Kong, Xiangtao and Qiao, Yu and Zhou, Jiantao and Dong, Chao},
  journal={IEEE Transactions on Pattern Analysis and Machine Intelligence},
  year={2025},
  publisher={IEEE}
}

@inproceedings{ntire26deepfake, 
title={{    Robust Deepfake Detection, NTIRE 2026 Challenge: Report    }}, 
author={    Hopf, Benedikt and  Timofte, Radu and others    },   
booktitle={Proceedings of the IEEE/CVF Conference on Computer Vision and Pattern Recognition (CVPR) Workshops},  
year = {2026} 
}

@inproceedings{ntire26hrdepth, 
title={{    NTIRE 2026 Challenge on High-Resolution Depth of non-Lambertian Surfaces    }}, 
author={    Zama Ramirez, Pierluigi and  Tosi, Fabio and  Di Stefano, Luigi and  Timofte, Radu and  Costanzino, Alex and  Poggi, Matteo and  Salti, Samuele and  Mattoccia, Stefano and others    },   
booktitle={Proceedings of the IEEE/CVF Conference on Computer Vision and Pattern Recognition (CVPR) Workshops},  
year = {2026} 
}

@inproceedings{ntire26raim_fusion, 
title={{    NTIRE 2026 The 3rd Restore Any Image Model (RAIM) Challenge: Multi-Exposure Image Fusion in Dynamic Scenes (Track2)    }}, 
author={    Qu, Lishen and  Liu, Yao and  Liang, Jie and  Zeng, Hui and  Dai, Wen and  Guan, Ya-nan and  Qin, Guanyi and  Zhou, Shihao and  Yang, Jufeng and  Zhang, Lei and  Timofte, Radu and others    },   
booktitle={Proceedings of the IEEE/CVF Conference on Computer Vision and Pattern Recognition (CVPR) Workshops},  
year = {2026} 
}

@inproceedings{ntire26raim_portrait, 
title={{    NTIRE 2026 The 3rd Restore Any Image Model (RAIM) Challenge: AI Flash Portrait (Track 3)    }}, 
author={    Guan, Ya-nan and  Zhang, Shaonan and  Guo, Hang and  Wang, Yawen and  Fan, Xinying and  Liang, Jie and  Zeng, Hui and  Qin, Guanyi and  Qu, Lishen and  Dai, Tao and  Xia, Shu-Tao and  Zhang, Lei and  Timofte, Radu and others    },   
booktitle={Proceedings of the IEEE/CVF Conference on Computer Vision and Pattern Recognition (CVPR) Workshops},  
year = {2026} 
}

@inproceedings{ntire26raim_piqa, 
title={{    NTIRE 2026 The 3rd Restore Any Image Model (RAIM) Challenge: Professional Image Quality Assessment (Track 1)    }}, 
author={    Qin, Guanyi and  Liang, Jie and  Zhang, Bingbing and  Qu, Lishen and  Guan, Ya-nan and  Zeng, Hui and  Zhang, Lei and  Timofte, Radu and others    },   
booktitle={Proceedings of the IEEE/CVF Conference on Computer Vision and Pattern Recognition (CVPR) Workshops},  
year = {2026} 
}

@inproceedings{ntire26lightsr, 
title={{    NTIRE 2026 Challenge on Light Field Image Super-Resolution: Methods and Results    }}, 
author={    Wang, Yingqian and  Liang, Zhengyu and  Zhang, Fengyuan and  Zhao, Wending and  Wang, Longguang and  Li, Juncheng and  Yang, Jungang and  Timofte, Radu and  Guo, Yulan and others    },   
booktitle={Proceedings of the IEEE/CVF Conference on Computer Vision and Pattern Recognition (CVPR) Workshops},  
year = {2026} 
}

@inproceedings{ntire263dsr, 
title={{    NTIRE 2026 Challenge on 3D Content Super-Resolution: Methods and Results    }}, 
author={    Wang, Longguang and  Guo, Yulan and  Wang, Yingqian and  Li, Juncheng and  Peng, Sida and  Zhang, Ye and  Timofte, Radu and  Chen, Minglin and  Wang, Yi and  Hu, Qibin and  Lei, Wenjie and others    },   
booktitle={Proceedings of the IEEE/CVF Conference on Computer Vision and Pattern Recognition (CVPR) Workshops},  
year = {2026} 
}

@inproceedings{ntire26videores, 
title={{    NTIRE 2026 Challenge on Bitstream-Corrupted Video Restoration: Methods and Results    }}, 
author={    Zou, Wenbin and  Liu, Tianyi and  Wu, Kejun and  Zhuang, Huiping and  Wu, Zongwei and  Zhou, Zhuyun and  Timofte, Radu and  others     },   booktitle={Proceedings of the IEEE/CVF Conference on Computer Vision and Pattern Recognition (CVPR) Workshops},  
year = {2026} 
}

@inproceedings{ntire26XAIGCqa, 
title={{    NTIRE 2026 X-AIGC Quality Assessment Challenge: Methods and Results    }}, 
author={    Liu, Xiaohong and  Min, Xiongkuo and  Zhai, Guangtao and  Hu, Qiang and  Cao, Jiezhang and  Zhou, Yu and  Sun, Wei and  Wen, Farong and  Xu, Zitong and  Zhou, Yingjie and  Duan, Huiyu and  Liu, Lu and  Wang, Jiarui and  Luo, Siqi and  Li, Chunyi and  Xu, Li and  Zhang, Zicheng and  Shi, Yue and  Wang, Yubo and  Zhang, Minghong and  Guo, Chunchao and  Hu, Zhichao and  Chen, Mingtao and  Wu, Xiele and  Ma, Xin and  Lv, Zhaohe and  Xue, Yuanhao and  Wang, Jiaqi and  Sha, Xinxing and  Timofte, Radu and  others    },   
booktitle={Proceedings of the IEEE/CVF Conference on Computer Vision and Pattern Recognition (CVPR) Workshops},  
year = {2026} 
}

@inproceedings{ntire26shadow, 
title={{    Advances in Single-Image Shadow Removal: Results from the NTIRE 2026 Challenge    }}, 
author={    Vasluianu, Florin-Alexandru and  Seizinger, Tim and  Zhou, Zhuyun and  Wu, Zongwei and  Timofte, Radu and  others     },   
booktitle={Proceedings of the IEEE/CVF Conference on Computer Vision and Pattern Recognition (CVPR) Workshops},  
year = {2026} 
}

@inproceedings{ntire26lightnorm, 
title={{    Learning-Based Ambient Lighting Normalization: NTIRE 2026 Challenge Results and Findings    }}, 
author={    Vasluianu, Florin-Alexandru and  Seizinger, Tim and  Chen, Jeffrey and  Zhou, Zhuyun and  Wu, Zongwei and  Timofte, Radu and  others    },   booktitle={Proceedings of the IEEE/CVF Conference on Computer Vision and Pattern Recognition (CVPR) Workshops},  
year = {2026} 
}

@inproceedings{ntire26bokeh, 
title={{    The First Controllable Bokeh Rendering Challenge at NTIRE 2026    }}, 
author={    Seizinger, Tim and  Vasluianu, Florin-Alexandru and  Conde, Marcos V. and  Chen, Jeffrey and  Zhou, Zhuyun and  Wu, Zongwei and  Timofte, Radu and  others    },   
booktitle={Proceedings of the IEEE/CVF Conference on Computer Vision and Pattern Recognition (CVPR) Workshops},  
year = {2026} 
}

@inproceedings{ntire26ripdetseg, 
title={{    NTIRE 2026 Rip Current Detection and Segmentation (RipDetSeg) Challenge Report    }}, 
author={    Dumitriu, Andrei and  Ralhan, Aakash and  Miron, Florin and  Tatui, Florin and  Ionescu, Radu Tudor and  Timofte, Radu and  others     },   booktitle={Proceedings of the IEEE/CVF Conference on Computer Vision and Pattern Recognition (CVPR) Workshops},  
year = {2026} 
}

@inproceedings{ntire26llie, 
title={{    Low Light Image Enhancement Challenge at NTIRE 2026    }}, 
author={    Ciubotariu, George and  S M A,  Sharif and  Rehman, Abdur and  Ali, Fayaz and  Naqvi, Rizwan Ali and  Conde, Marcos and  Timofte, Radu and others    },   
booktitle={Proceedings of the IEEE/CVF Conference on Computer Vision and Pattern Recognition (CVPR) Workshops},  
year = {2026} 
}

@inproceedings{ntire26highfps, 
title={{    High FPS Video Frame Interpolation Challenge at NTIRE 2026    }}, 
author={    Ciubotariu, George and  Zhou, Zhuyun and  Jin, Yeying and  Wu, Zongwei and  Timofte, Radu and  others    },   
booktitle={Proceedings of the IEEE/CVF Conference on Computer Vision and Pattern Recognition (CVPR) Workshops},  
year = {2026} 
}

@inproceedings{ntire26nthaze, 
title={{    NT-HAZE: A Benchmark Dataset for Realistic Night-time Image Dehazing    }}, 
author={    Ancuti, Radu and  Ancuti, Codruta and  Timofte, Radu and  Ancuti, Cosmin    },   
booktitle={Proceedings of the IEEE/CVF Conference on Computer Vision and Pattern Recognition (CVPR) Workshops},  
year = {2026} 
}

@inproceedings{ntire26nthaze_rep, 
title={{    NTIRE 2026 Nighttime Image Dehazing Challenge Report    }}, 
author={    Ancuti, Radu and  Brateanu, Alexandru and  Vasluianu, Florin and  Balmez, Raul and  Orhei, Ciprian and  Ancuti, Codruta and  Timofte, Radu and  Ancuti, Cosmin and others    },   
booktitle={Proceedings of the IEEE/CVF Conference on Computer Vision and Pattern Recognition (CVPR) Workshops},  
year = {2026} 
}

@inproceedings{ntire26isp, 
title={{    NTIRE 2026 Challenge on Learned Smartphone ISP with Unpaired Data: Methods and Results    }}, 
author={    Perevozchikov, Georgy and  Vladimirov, Daniil and  Timofte, Radu and  others    },   
booktitle={Proceedings of the IEEE/CVF Conference on Computer Vision and Pattern Recognition (CVPR) Workshops},  
year = {2026} 
}

@inproceedings{ntire26ugcvideo, 
title={{    NTIRE 2026 Challenge on Short-form UGC Video Restoration in the Wild with Generative Models: Datasets, Methods and Results    }}, author={    Li, Xin and  Gong, Jiachao and  Wang, Xijun and  Xiong, Shiyao and  Li, Bingchen and  Yao, Suhang  and  Zhou, Chao and  Chen, Zhibo and  Timofte, Radu and others    },   
booktitle={Proceedings of the IEEE/CVF Conference on Computer Vision and Pattern Recognition (CVPR) Workshops},  
year = {2026} 
}

@inproceedings{ntire26dual_focus, 
title={{    NTIRE 2026 The Second Challenge on Day and Night Raindrop Removal for Dual-Focused Images: Methods and Results    }}, 
author={    Li, Xin and  Jin, Yeying and  Yao, Suhang and  Lin, Beibei and  Fan, Zhaoxin and   Yan, Wending and  Jin, Xin and  Wu, Zongwei  and  Li, Bingchen  and  Shi, Peishu and  Yang, Yufei and  Li, Yu and  Chen, Zhibo  and  Wen, Bihan and  Tan, Robby and  Timofte, Radu and others    },   
booktitle={Proceedings of the IEEE/CVF Conference on Computer Vision and Pattern Recognition (CVPR) Workshops},  
year = {2026} 
}

@inproceedings{ntire26srx4, 
title={{    The Fourth Challenge on Image Super-Resolution (×4) at NTIRE 2026: Benchmark Results and Method Overview    }}, 
author={    Chen, Zheng and  Liu, Kai and  Wang, Jingkai and  Yan, Xianglong and  Li, Jianze and  Zhang, Ziqing and  Gong, Jue and  Li, Jiatong and  Sun, Lei and  Liu, Xiaoyang and  Timofte, Radu and  Zhang, Yulun and others    },   
booktitle={Proceedings of the IEEE/CVF Conference on Computer Vision and Pattern Recognition (CVPR) Workshops},  
year = {2026} 
}

@inproceedings{ntire26retouching, 
title={{    Photography Retouching Transfer, NTIRE 2026 Challenge: Report    }}, 
author={    Elezabi, Omar and  V. Conde, Marcos and  Wu, Zongwei and  Jin, Yeying and  Timofte, Radu and others    },   
booktitle={Proceedings of the IEEE/CVF Conference on Computer Vision and Pattern Recognition (CVPR) Workshops},  
year = {2026} 
}

@inproceedings{ntire26rwsr, 
title={{    The First Challenge on Mobile Real-World Image Super-Resolution at NTIRE 2026: Benchmark Results and Method Overview    }}, 
author={    Li, Jiatong and  Chen, Zheng and  Liu, Kai and  Wang, Jingkai and  Zhou, Zihan and  Liu, Xiaoyang and  Zhu, Libo and  Timofte, Radu and  Zhang, Yulun and others    },   
booktitle={Proceedings of the IEEE/CVF Conference on Computer Vision and Pattern Recognition (CVPR) Workshops},  
year = {2026} 
}

@inproceedings{ntire26rsirsr, 
title={{    The First Challenge on Remote Sensing Infrared Image Super-Resolution at NTIRE 2026: Benchmark Results and Method Overview    }}, author={    Liu, Kai and  Yue, Haoyang and  Lin, Zeli and  Chen, Zheng and  Wang, Jingkai and  Gong, Jue and  Timofte, Radu and  Zhang, Yulun and  others    },   
booktitle={Proceedings of the IEEE/CVF Conference on Computer Vision and Pattern Recognition (CVPR) Workshops},  
year = {2026} 
}

@inproceedings{ntire26aigendet, 
title={{    NTIRE 2026 Challenge on Robust AI-Generated Image Detection in the Wild    }}, 
author={    Gushchin, Aleksandr and  Abud, Khaled and  Shumitskaya, Ekaterina and  Filippov, Artem and  Bychkov, Georgii and  Lavrushkin, Sergey and  Erofeev, Mikhail and  Antsiferova, Anastasia and  Chen, Changsheng and  Tan, Shunquan and  Timofte, Radu and  Vatolin, Dmitriy and others    },
booktitle={Proceedings of the IEEE/CVF Conference on Computer Vision and Pattern Recognition (CVPR) Workshops},  
year = {2026} 
}

@inproceedings{ntire26cdfsod, 
title={{    The Second Challenge on Cross-Domain Few-Shot Object Detection at NTIRE 2026: Methods and Results    }}, 
author={    Qiu, Xingyu and  Fu, Yuqian and  Geng, Jiawei and  Ren, Bin and  Pan, Jiancheng and  Wu, Zongwei and  Tang, Hao and  Fu, Yanwei and  Timofte, Radu and  Sebe, Nicu and  Elhoseiny, Mohamed and others    },   
booktitle={Proceedings of the IEEE/CVF Conference on Computer Vision and Pattern Recognition (CVPR) Workshops},  
year = {2026} 
}

@inproceedings{ntire26finrec, 
title={{    NTIRE 2026 Challenge on End-to-End Financial Receipt Restoration and Reasoning from Degraded Images: Datasets, Methods and Results    }}, author={    Guan, Bochen and  Li, Jinlong and  Yang, Kangning and  Ke, Chuang and  Cai, Jie and  Vasluianu, Florin and  Timofte, Radu and others    },   booktitle={Proceedings of the IEEE/CVF Conference on Computer Vision and Pattern Recognition (CVPR) Workshops},  
year = {2026} 
}

@inproceedings{ntire26faceres, 
title={{    The Second Challenge on Real-World Face Restoration at NTIRE 2026: Methods and Results    }}, 
author={    Wang, Jingkai and  Gong, Jue and  Chen, Zheng and  Liu, Kai and  Li, Jiatong and  Zhang, Yulun and  Timofte, Radu and  others    },
booktitle={Proceedings of the IEEE/CVF Conference on Computer Vision and Pattern Recognition (CVPR) Workshops},  
year = {2026} 
}

@inproceedings{ntire26reflection, 
title={{    NTIRE 2026 Challenge on Single Image Reflection Removal in the Wild: Datasets, Results, and Methods    }}, 
author={    Cai, Jie and  Yang, Kangning and  Li, Zhiyuan and  Vasluianu, Florin and  Timofte, Radu and others    },   
booktitle={Proceedings of the IEEE/CVF Conference on Computer Vision and Pattern Recognition (CVPR) Workshops},  
year = {2026} 
}

@inproceedings{ntire26anomalydet, 
title={{    NTIRE 2026  Challenge Report on Anomaly Detection of Face Enhancement for UGC Images    }}, 
author={    Zhong, Yan and   Ma,  Qiufang and  Wang, Zhen and  Jiang, Tingting and  Timofte, Radu and others    },   
booktitle={Proceedings of the IEEE/CVF Conference on Computer Vision and Pattern Recognition (CVPR) Workshops},  
year = {2026} 
}

@inproceedings{ntire26videosal, 
title={{    NTIRE 2026 Challenge on Video Saliency Prediction: Methods and Results    }}, 
author={    Moskalenko, Andrey and  Bryncev, Alexey and  Kosmynin, Ivan and  Shilovskaya, Kira and  Erofeev, Mikhail and  Vatolin, Dmitry and  Timofte, Radu and others    },   
booktitle={Proceedings of the IEEE/CVF Conference on Computer Vision and Pattern Recognition (CVPR) Workshops},  
year = {2026} 
}

@inproceedings{ntire26effsr, 
title={{    The Eleventh NTIRE 2026 Efficient Super-Resolution Challenge Report    }}, 
author={    Ren, Bin and  Guo, Hang and  Shu, Yan and  Ma, Jiaqi and  Cui, Ziteng and  Liu, Shuhong  and  Mei, Guofeng  and  Sun, Lei and  Wu, Zongwei and  Khan, Fahad Shahbaz and  Khan, Salman and  Timofte, Radu and  Li, Yawei and others    },   
booktitle={Proceedings of the IEEE/CVF Conference on Computer Vision and Pattern Recognition (CVPR) Workshops},  
year = {2026} 
}

@inproceedings{ntire26realx3d, 
title={{    3D Restoration and Reconstruction in Adverse Conditions: RealX3D Challenge Results    }}, 
author={    Liu, Shuhong and  Cui, Ziteng and  Bao, Chenyu and  Chu, Xuangeng and  Gu, Lin and  Ren, Bin and  Timofte, Radu and  Conde, Marcos V. and others    },   
booktitle={Proceedings of the IEEE/CVF Conference on Computer Vision and Pattern Recognition (CVPR) Workshops},  
year = {2026} 
}

@inproceedings{ntire26denoising, 
title={{    The Third Challenge on Image Denoising at NTIRE 2026: Methods and Results    }}, 
author={    Sun, Lei and  Guo, Hang and  Ren, Bin and  Su, Shaolin and  Wang, Xian and  Pani Paudel, Danda and  Van Gool, Luc and  Timofte, Radu and  Li, Yawei and others    },   
booktitle={Proceedings of the IEEE/CVF Conference on Computer Vision and Pattern Recognition (CVPR) Workshops},  
year = {2026} 
}

@inproceedings{ntire26aberration, 
title={{    NTIRE 2026 The First Challenge on Blind Computational Aberration Correction: Methods and Results    }}, 
author={    Sun, Lei and  Qian, Xiaolong and  Jiang, Qi and  Wang, Xian and  Gao, Yao and  Yang, Kailun and  Wang, Kaiwei and  Timofte, Radu and  Pani Paudel, Danda and  Van Gool, Luc and others    },   
booktitle={Proceedings of the IEEE/CVF Conference on Computer Vision and Pattern Recognition (CVPR) Workshops},  
year = {2026} 
}

@inproceedings{ntire26eventblurr, 
title={{    The Second Challenge on Event-Based Image Deblurring at NTIRE 2026: Methods and Results    }}, 
author={    Sun, Lei and  Li, Weilun and  Wang, Xian and  Li, Zhendong and  Shi, Letian and  Xu, Dannong and  Zhang, Deheng and  Hu, Mengshun and  Guo, Shuang and  Su, Shaolin and  Timofte, Radu and  Pani Paudel, Danda and  Van Gool, Luc and others    },   
booktitle={Proceedings of the IEEE/CVF Conference on Computer Vision and Pattern Recognition (CVPR) Workshops},  
year = {2026} 
}

@inproceedings{ntire26bursthdr, 
title={{    NTIRE 2026 Challenge on Efficient Burst HDR and Restoration: Datasets, Methods, and Results    }}, 
author={    Park, Hyunhee and  Park, Eunpil and  Lee, Sangmin and  Timofte, Radu and others    },   
booktitle={Proceedings of the IEEE/CVF Conference on Computer Vision and Pattern Recognition (CVPR) Workshops},  
year = {2026} 
}

@inproceedings{ntire26twilight, 
title={{    NTIRE 2026 Low-light Enhancement: Twilight Cowboy Challenge    }}, 
author={    Khalin, Aleksei and  Ershov, Egor and  Panshin, Artem and  Korchagin, Sergey and  Lobarev, Georgiy and  Terekhin, Arseniy and  Dorogova, Sofiia and  Shamsutdinov, Amir and  Mamedov, Yasin and  Khalfin, Bakhtiyar and  Sheludko, Bogdan and  Zilyaev, Emil and  Banić, Nikola and  Perevozchikov, Georgy and  Timofte, Radu and others    },   
booktitle={Proceedings of the IEEE/CVF Conference on Computer Vision and Pattern Recognition (CVPR) Workshops},  
year = {2026} 
}

@inproceedings{ntire26effllie, 
title={{    Efficient Low Light Image Enhancement: NTIRE 2026 Challenge Report    }}, 
author={    Yan, Jiebin  and  Tu, Chenyu  and  Lin, Qinghua and  WU, Zongwei and  Zhang , Weixia and  Wang, Zhihua and  Cao, Peibei and  Fang, Yuming  and  Liu, Xiaoning  and  Zhou, Zhuyun and  Timofte, Radu  and  others    },   
booktitle={Proceedings of the IEEE/CVF Conference on Computer Vision and Pattern Recognition (CVPR) Workshops},  
year = {2026} 
}

@inproceedings{timofte2016seven,
  title={Seven ways to improve example-based single image super resolution},
  author={Timofte, Radu and Rothe, Rasmus and Van Gool, Luc},
  booktitle={Proceedings of the IEEE conference on computer vision and pattern recognition},
  pages={1865--1873},
  year={2016}
}
}

\end{document}